\documentclass[review]{elsarticle}

\usepackage{amssymb}
\usepackage{arabtex}
\usepackage{url}
\usepackage{utf8}
\usepackage[utf8x]{inputenc}
\usepackage[T1]{fontenc}
\usepackage{subfig}
\usepackage{algorithm2e}
\usepackage{longtable}
\usepackage{multirow, pbox}
\novocalize

\journal{Neurocomputing}
\begin{document}

\begin{frontmatter}

\title{A Novel Context-Aware Multimodal Framework for Persian Sentiment Analysis}

\author[label1]{Kia Dashtipour}
\address[label1]{Department of Computing Science and Mathematics, University of Stirling, Stirling, UK}

\author[label4]{Mandar Gogate}
\cortext[cor1]{Corresponding author: Mandar Gogate}
\ead{M.Gogate@napier.ac.uk}
\address[label4]{School of Computing, Edinburgh Napier University, Edinburgh, UK}

\author[label6]{Erik Cambria}
\address[label6]{School of Computer Science and Engineering, Nanyang Technological University, Singapore}

\author[label4]{Amir Hussain}

\begin{abstract}

Most recent works on sentiment analysis have exploited the text modality. However, millions of hours of video recordings posted on social media platforms everyday hold vital unstructured information that can be exploited to more effectively gauge public perception. Multimodal sentiment analysis offers an innovative solution to computationally understand and harvest sentiments from videos by contextually exploiting audio, visual and textual cues. In this paper, we, firstly, present a first of its kind Persian multimodal dataset comprising more than 800 utterances, as a benchmark resource for researchers to evaluate multimodal sentiment analysis approaches in Persian language. Secondly, we present a novel context-aware multimodal sentiment analysis framework, that simultaneously exploits acoustic, visual and textual cues to more accurately determine the expressed sentiment. We employ both decision-level (late) and feature-level (early) fusion methods to integrate affective cross-modal information. 
Experimental results demonstrate that the contextual integration of multimodal features such as textual, acoustic and visual features deliver better performance (91.39\%) compared to unimodal features (89.24\%).
}
\end{abstract}

\begin{keyword}
\texttt{Multimodal Sentiment Analysis \sep Persian Sentiment Analysis}
\end{keyword}

\end{frontmatter}

\section{Introduction}

The emergence of online forums has increased exponentially in the last few years~\cite{cambig}. Furthermore, with the advent of social media (e.g., Twitter, YouTube and Facebook), people can share their opinions frequently. Social media encourage people to engage in discussions and enables them to share their thoughts on a range of issues and challenges~\cite{grasen}. Online media provide a platform for sharing ideas and encourages public to join group discussions. In addition, social media allow companies and organizations to get feedback regarding their products in the form of texts, images and videos~\cite{rabby2020teket,camben,rabby2020teket,zhong2020extracting,satrev,angbri,raglea}. 

\begin{figure}[!t]
\centering
\includegraphics[width=\textwidth]{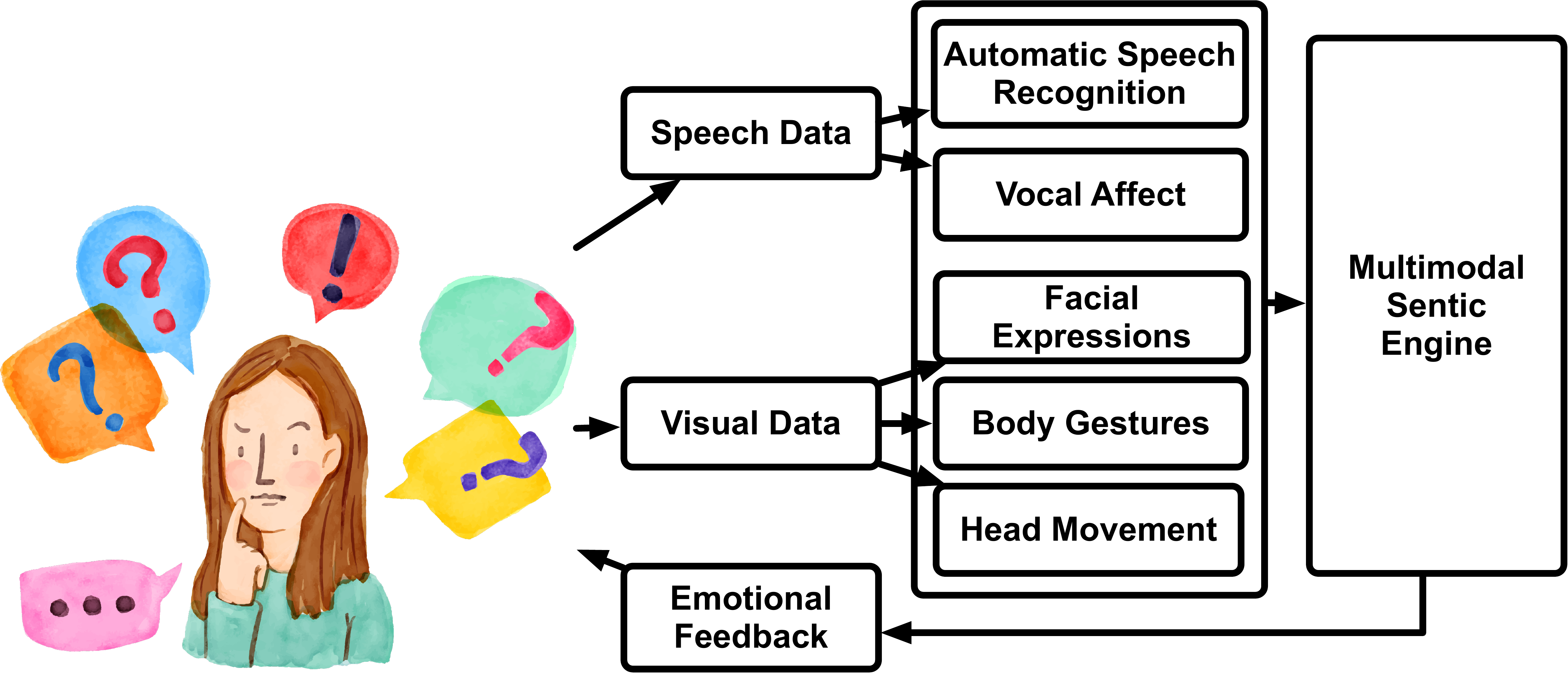}
\caption {A generic multimodal sentiment analysis framework~\cite{poria2016fusing}}
\label{fig:irdfpo}
\end{figure}

To date, most current research in sentiment analysis has been conducted on textual data \cite{dashtipour2018exploiting,dashtipour2017persian,dashtipour2017comparative,hussain2020artificial}. As a result, most developed resources are limited to sentiment analysis from text. However, with advent of social media, people are extensively using social media to express their opinions in different languages~\cite{jones1807grammar,loomul,ahmed2019offline}. In addition, they are increasingly using videos, audio, images and text to express their opinions in social media. Therefore, it is becoming vital to identify sentiment in various modalities as well as in different languages~\cite{yadav2015multimodal, jiangrobust}. An overview of a typical multimodal sentiment analysis framework is shown in Fig.~\ref{fig:irdfpo}, which follows our pioneering works in~\cite{chaturvedi2019fuzzy,camble,traens,camalb,porens,cambria2020senticnet}. As shown in Fig.~\ref{fig:irdfpo}, the visual, acoustic and textual features are extracted to determines the overall polarity of the input videos. First, the video is transcribed into text, then, the acoustic features are extracted using openSMILE and finally visual features extracted and all features concatenated to find overall polarity of the sentence.

Persian comprises 32 letters, which cover 28 Arabic letters. Its writing system includes special signs and diacritic marks that can be used in different forms or omitted from the word. The Persian language has more than 100 million speakers in Iran, Afghanistan and Tajikistan. Key challenges in Persian language are sarcasm, idioms, use of informal words, phrases and more spelling mistakes compared to other languages such as English~\cite{dashtipour2016persent,dashtipour2019persent,dashtipour2020ensemble}. In our recent work~\cite{dashtipour2020hybrid}, we proposed a first of its kind, linguistic dependency-rules (pattern) based deep learning approach for Persian sentiment analysis.

In this paper, we address the more challenging task of Persian multimodal sentiment analysis, by developing a novel context-aware framework and conduct an extensive set of experiments to show that a multimodal model that contextually combines visual, acoustic and text features can significantly enhance the sentiment polarity detection of online Persian videos. 

The main aim of this paper is to address the task of multimodal sentiment analysis by conducting a proof-of-concept set of experiments to show that the multimodal features including visual, acoustic and textual features in Persian language can be more effective to identify the overall sentiment of multimodal data. Therefore, in this paper, we address the first-time task for tri-modal sentiment analysis by integrating three different modalities (audio, visual and text) which are integrated to identify the polarity of the input data. This is unlike the previous studies on sentiment analysis which only considers text modality. Second, we presented a first of its kind novel Persian multimodal sentiment analysis data collected from YouTube to enable training of multimodal sentiment analysis models. The experimental results show that the multimodal features (textual, acoustic and visual) achieved better performance as compared unimodal features.

In summary, the paper reports four major contributions outlined below:

\begin{enumerate}
\item We propose a novel context-aware multimodal fusion framework to extract unimodal and multimodal features using automated feature extraction methods including convolutional neural network (CNN) and long short-term memory (LSTM). The features are integrated to perform multimodal sentiment analysis task.

\item A first of its kind Persian multimodal dataset is presented, consisting of video opinions, collected from YouTube, which is analyzed and annotated to enable multimodal sentiment analysis.

\item We conduct an extensive set of experiments to demonstrate that a multimodal approach contextually exploiting visual, acoustic and textual features can more accurately determine the sentiment polarity of online Persian videos, compared to state-of-the-art unimodal approaches (e.g. text-based).

\item The context-aware multimodal approach is shown to overcome the limitations of ambiguous words usage in Persian. For example, Persian textual reviews often comprise ambiguous words and phrases such as \RL{fylm aft.dA.h xwb bwd}, `The movie is awfully good' which can lead to incorrect classification of data. In such cases, the contextual use of acoustic and visual features enable more accurate polarity assignment through the different modalities. 

\end{enumerate}

The rest of the paper is organized as follows: Section 2 presents related work on state-of-the-art multimodal sentiment analysis approaches for English and other languages; Section~\ref{multi:methodology} presents the proposed framework for Persian multimodal sentiment analysis; Section~\ref{multi:multidataset} presents the novel Persian multimodal dataset; Section~\ref{multi:experiment} presents comparative experimental results; finally, Section~\ref{multi:conclusion} concludes this paper and outlines some future work directions.

\section{Related Work}

In the literature, extensive research has been carried out to develop multimodal models, that demonstrate the significance of multimodal processing over unimodal approaches. Researchers have proposed multimodal processing architectures for a wide range of real-world applications ranging from sentiment analysis~\cite{tahir2020novel, jiang2020densely,elayeb2020automatic,cambio,adeel2019survey, jiang2020robust,asad2020travelers,yu2020energy}, deception detection~\cite{gogate2017deep} to dementia diagnosis and progression prediction~\cite{ieracitano2020novel} and audio-visual speech recognition and enhancement~\cite{hu2019deep,gogate2020cochleanet,adeel2019lip,howard2019deep}. 

In related work, Morency~\cite{morency2011towards} presented an approach for Spanish multimodal sentiment analysis using the MOUD dataset which is product reviews videos collected from YouTube to evaluate its performance. Comparative experimental results showed that the combination of audio, visual and text features can improve the performance of the approach. Poria et al.~\cite{poria2015deep} presented a novel method to extract text features for short text as part of a multimodal sentiment analysis framework. Their dataset comprised 498 (in English) short videos and fused audio, text and visual features were used to train their model.

Poria et al.~\cite{poria2016fusing} also proposed a method for English multimodal sentiment analysis to identify sentiments using different approaches to fuse audio, text and video features. Their proposed method was evaluated with a YouTube dataset, however it was restricted to a single speaker and was unable to identify the overall polarity of video with multiple speakers. On the other hand, Rosas et al.~\cite{rosas2013multimodal} proposed an approach for Spanish multimodal sentiment analysis. Due to lack of resources, Spanish videos were collected from YouTube to enable audio, video and textual features extraction. A support vector machine (SVM) was used to evaluate the performance of the approach. Experimental results showed that the combination of text, audio and video features achieved better performance as compared to text and audio modalities. Their approach identified the positive or negative polarity of sentence but was unable to detect the polarity for neutral sentences.

Alqarafi et al.~\cite{alqarafi2017toward} proposed a multimodal approach to detect polarity in Arabic videos. A total of 40 videos in Arabic language were collected from YouTube and manually transcribed. It is worth to mention that, the collected videos is only cover Arabic Saudi accent and it cannot generalized on different Arabic dialects.   
In order to evaluate the performance of their approach, text features such as ngram and visual features (smile, frown, head nod, and head shake) were extracted. Empirical results demonstrated that the combination of visual and text features was more effective as compared to unimodal features. Their approach did not however consider acoustic features and only relied on text and visual features. 

Dastgheib et al.~\cite{dastgheib2020application} proposed a novel hybrid method using a combination of feature-based and deep convolutional neural network (CNN) to detect polarity in Persian movie reviews. The results show the proposed hybrid method displayed the combination of feature-based and CNN achieved better performance as compared to CNN and LSTM. Farahani et al.~\cite{farahani2020parsbert}  used a BERT architecture to develop a pre-trained framework to detect polarity in Persian product reviews, which is called ParsBERT. The model is based on different NLP tasks such as named entity recognition and sentiment analysis. The experimental results show that the proposed architecture achieved better performance as compared to multilingual approaches.

However, none of the aforementioned works have explored polarity detection challenges for Persian multimodal sentiment analysis. Motivated by this, in this paper we propose a novel framework coupled with a first of its kind multimodal Persian dataset, which are described in Section~\ref{multi:multidataset}. Table 1 shows the summary of the proposed approach in multimodal sentiment analysis.

\begin{table}[]
\label{tab:12sdsd}
\caption{Summary of the multimodal sentiment analysis approaches}
\begin{tabular}{|l|l|l|l|}
\hline
\textbf{Ref}      & \textbf{Method}               & \textbf{Dataset}                                                                        & \textbf{Accuracy} \\ \hline
Alqarafi et al. \cite{alqarafi2017toward}  & Arabic Multimodal             & \begin{tabular}[c]{@{}l@{}}YouTube Videoes\\ Arabic Saudi dialects\end{tabular}         & 65.59             \\ \hline
Dashtipour et al. \cite{dashtipour2020hybrid} & Hybrid Rule-based             & \begin{tabular}[c]{@{}l@{}}Persian movie review\\ Only text modality\end{tabular}       & 86.29             \\ \hline
Dastgheib et al. \cite{dastgheib2020application}  & Hybrid Approach               & \begin{tabular}[c]{@{}l@{}}Persian product reviews\\ Only text modality\end{tabular}    & 74                \\ \hline
Farahani et al. \cite{farahani2020parsbert}  & BERT Method                   & \begin{tabular}[c]{@{}l@{}}Persian product reviews\\ Only textual modality\end{tabular} & 98.79             \\ \hline
Poria et al.\cite{poria2016fusing}      & Multimodal SA & English YouTube videos                                                                  & 77.01             \\ \hline
Rosas et al  \cite{rosas2013multimodal}     & Multimodal SA & YouTube videos Spanish                                                                  & 75                \\ \hline
\end{tabular}
\end{table}

\section{Methodology} \label{multi:methodology}

\begin{figure}[!t]
\centering
\includegraphics[width=\textwidth]{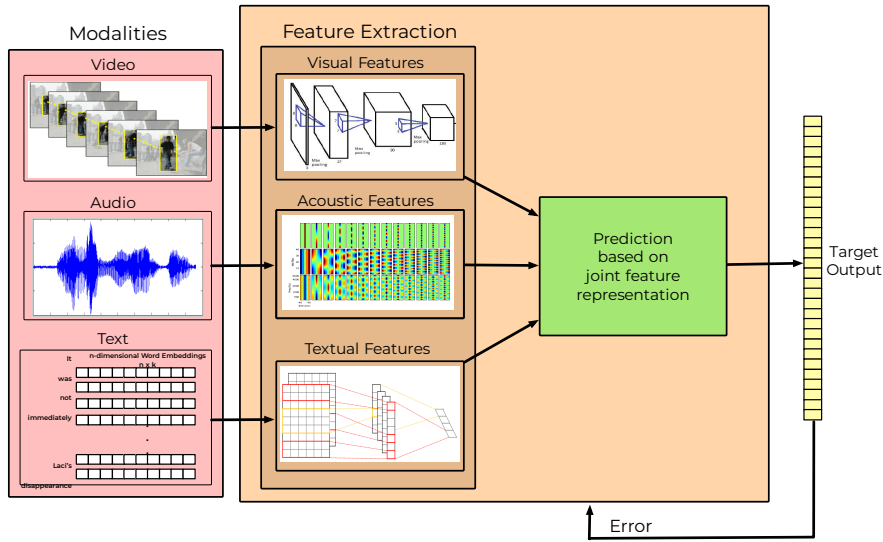}
\caption {Overview of the Persian Multimodal Sentiment Analysis Framework}
\label{fig:future}
\end{figure}

In this section, the proposed multimodal Persian sentiment analysis framework, shown in Fig.~\ref{fig:future}, is described. As can be seen, audio, visual and textual features are first contextually extracted, and the extracted features, representing affective information, are then fused to identify the overall polarity of the target video dataset. 

\subsection{Unimodal Feature Extraction}

In this section, we discuss unimodal feature extraction techniques.

\subsubsection{Textual Feature Extraction: text-BiLSTM}
\begin{figure}[!t]
\centering
\label{LSTM:TEXT}
\includegraphics[width=\textwidth]{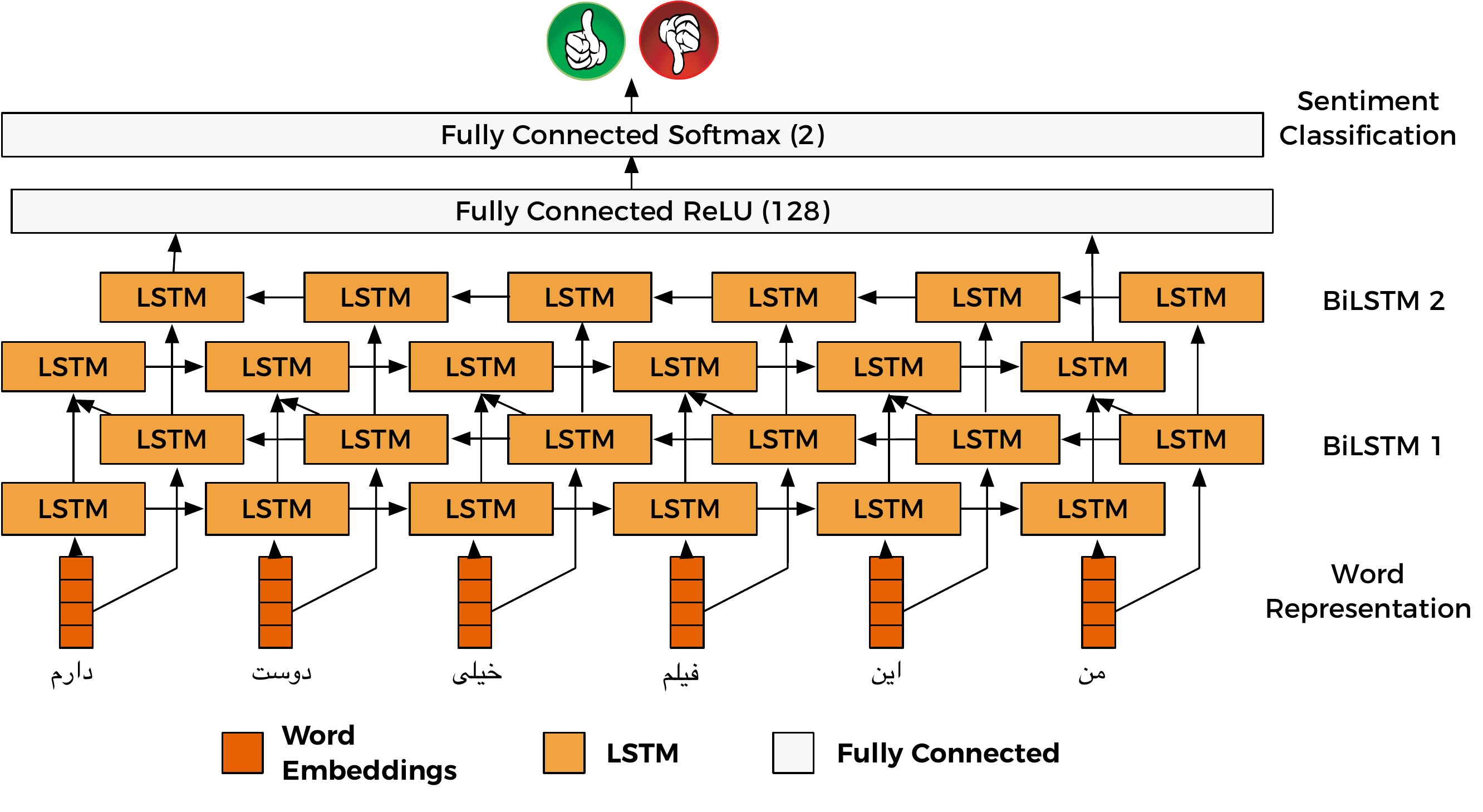}
\caption {Bidirectional LSTM architecture for Textual-cues feature extraction}
\label{fig:aud}
\end{figure}

For contextually extracting features from the textual modality, a stacked bidirectional LSTM (BiLSTM) model, as shown in Fig.~\ref{LSTM:TEXT} is used. Each utterance is represented as a concatenation of 300 dimensional pretrained fastText word embeddings. Each utterance is either trimmed with a window of size 60 words or zero padded at the end to form a vector of dimension 60 x 300. The converted vectors are fed into a stacked BiLSTM model, whose parameters are experimentally determined and optimized using a trial and error approach. The model consists of two bidirectional LSTM layers with 128 cells each. The output of the last bidirectional LSTM is concatenated and fed to a fully connected layer with 128 neurons (ReLU activation) and 2 neurons (Softmax activation) respectively. The network learn the levels of abstract representations and implicit semantic information, that spans over the entire utterance. 
The proposed implemented BiLSTM architecture is similar to the one which is used by Wang et al. ~\cite{wang2016attention}, contains of an input layer, two stacked LSTM layers along with an output fully connected layer. Particularly, the BiLSTM consists of two stacked bidirectional LSTM with 128 and 64 cells and dropout of 0.2 probability and dense layer with two neurons and softmax activation.

Following our recent work, we exploit a novel linguistic dependency rules (patterns) and deep learning based approach~\cite{dashtipour2020hybrid}, for the transcribed Persian utterances. As can be seen from example results in Table~\ref{multi:textsentic}, a deep learning based classifier can be used to effectively classify the unclassified Persian sentences, specifically sentences that cannot be categorized by dependency-based rules. For the results shown in Table~\ref{multi:textsentic}, each sentence is converted into a 300-dimensional vector using fastText and the concatenation of word embedding is fed into deep learning classifiers. For comparison, the sentences are converted into Bag-of-words and fed into logistic regression and SVM. In addition, the transcribed videos are converted into 300 dimensional word embeddings~\cite{joulin2016fasttext}, which are then fed into deep learning (CNN and LSTM) classifiers, with results shown in the Table 2.

The dataset used for the experiments divided into 60\% train set, 10\% validation set and 30\% testing set. The CNN, LSTM, SVM and LR were trained on the train set, tuned on validation set and evaluated on the test set. Comparative experimental results show that the hybrid 2: LSTM + Dependency-based achieved better accuracy as compared with other approaches including DNN based classifiers. The number of neutral discourses are very less as compared to positive and negative cases. Therefore, we cannot train our proposed models on neutral comments. As deep learning requires a large number of training and testing set.

\begin{table}[]
\caption{Textual Features Results}
\label{multi:textsentic}
\centering
\begin{tabular}{|l|l|l|l|l|}
\hline
Classifier                                  & Precision & Recall & F-measure & Accuracy \\ \hline
Kim et al.~\cite{kim2014convolutional}           & 0.61   & 0.63  & 0.61   & 61.24  \\ \hline
Dehkharghani et al.~\cite{dehkharghani2012adaptation}    & 0.78   & 0.82  & 0.79   & 70.12  \\ \hline
fastText Classifier~\cite{joulin2017bag}          & 0.67   & 0.67  & 0.67   & 70.01  \\ \hline
SVM                                     & 0.65   & 0.65  & 0.65   & 65.01  \\ \hline
Logistic Regression                             & 0.64   & 0.64  & 0.64   & 64.23  \\ \hline
Dependency-based Approach                          & 0.83   & 0.93  & 0.87   & 75.94  \\ \hline
CNN                                     & 0.91   & 0.63  & 0.75   & 68.53  \\ \hline
LSTM                                     & 0.92   & 0.83  & 0.87   & 86.14  \\ \hline
\begin{tabular}[c]{@{}l@{}}Hybrid 1: CNN + \\ Dependency-Based\end{tabular} & 0.78   & 0.77  & 0.77   & 76.16  \\ \hline
\begin{tabular}[c]{@{}l@{}}Hybrid 2: LSTM + \\ Dependency-Based\end{tabular} & 0.86   & 0.95  & 0.85   & 88.01  \\ \hline
\end{tabular}
\end{table}

\subsection{Audio Feature Extraction: openSMILE}

The current literature shows that the openSMILE can effectively extract audio features such as deception as well as sentiment \cite{soleymani2017survey,poria2015deep}. Therefore, the openSMILE is used because it can automatically extract low-level descriptors such as beat histogram, Mel frequency cepstral coefficients, spectral centroid, spectral flux, beat histogram, beat sum.

The audio features are automatically extracted from the speech of each utterance using the widely used OpenSMILE software. The openSMILE is used to extract features which consists of several low level descriptors and their statistical information. In addition, features such as amplitude mean, arithmetic, mean, quadratic mean, standard deviation, flatness, skewness, kurtosis, quartiles are also extracted. In total, we have obtained more than 6,373 features. The features are extracted at a frequency of 40 samples per second. The extracted features consist of the following acoustic sub-features: 

\begin{itemize}
 \item Prosody feature: This feature consists of intensity, loudness and pitch that describe the speech signal in terms of amplitude and frequency.
 \item Energy features: The energy feature depicts the human loudness perception.
 \item Voice probabilities: The voice probabilities provide an estimate of percentage of voiced and unvoiced energy in the audio.
 \item Spectral features: The spectral features are based on the characteristics of the human ear, which uses a nonlinear frequency unit to simulate the auditory system
 \item Cepstral features: The cepstral features emphasize the changes in the spectrum features measured by frequencies. Specifically, 12-model Mel-frequency cepstral coefficients are used and calculated based on the Fourier transform of a speech fame.
\end{itemize}

The overall audio features for a single utterance consist of 6373 features. Speaker normalization is performed using z-standardization. The voice intensity is a threshold to identify the samples with and without speech. The features are averaged over all the frames in an utterance, to obtain a feature vector for each utterance. The audio feature extraction framework is shown in Fig.~\ref{fig:aud}. The multilayer perceptron (MLP) architecture, as depicted in Table~\ref{tab:sixonek}, is used to exploit the extracted audio feature for determining the opinion strength based on acoustic cues. The MLP architecture is shown in Fig.~\ref{fig:aud}.

\begin{table}[!t]
\caption{MLP Architecture for audio feature extraction}
\label{tab:sixonek}
\centering
\begin{tabular}{|l|l|l|l|l|}
\hline
\textbf{Layer} & 1 & 2 & 3 & 4 \\ \hline
\textbf{Type} & Relu & ReLU & ReLU & ReLU \\ \hline
\textbf{Neurons} & 1024 & 512 & 128 & 1 \\ \hline
\end{tabular}
\end{table}

\begin{figure}[!t]
\centering
\includegraphics[width=\textwidth]{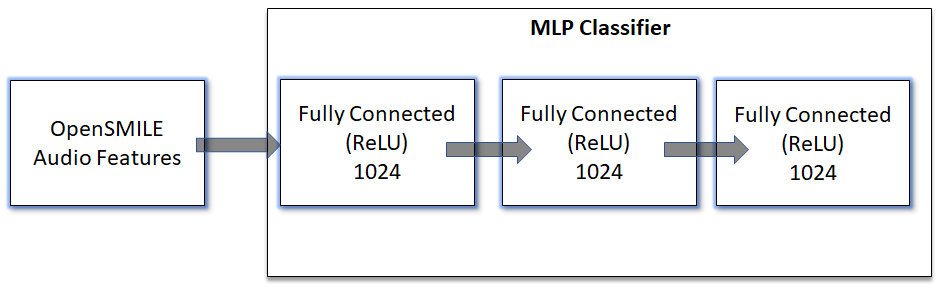}
\caption {Acoustic cues feature extraction}
\label{fig:aud}
\end{figure}

\subsection{Visual Feature Extraction: 3D-CNN}
\label{multi:visual}

Human expressions play a significant role in identifying the emotion expressed in day-to-day conversations~\cite{wanrev,sushou}. In particular, facial expressions help in decoding the expressed affect by providing visual cues. Therefore, visual features are important in multimodal sentiment analysis. In this work, a Facial Action Coding System (FACS) is used for measuring and describing facial behaviors. According to~\cite{turk1991face}, facial behavior can be categorized into 64 action units. The Computer Expression Recognition Toolbox (CERT) is employed to automatically extract the following visual features:

\textbf{Smile and head pose estimates}: The smile feature depicts the probability of a person smiling given an image. Head pose detection consists of three dimensional head orientation yaw, pitch and roll. These dimensions provide information about the face position while expressing positive or negative opinions~\cite{werner2017landmark}.

\textbf{Facial action units}: The facial action units estimate the thirty related muscle movements related to eyes, nose, eyebrows and chin. This feature provides information about facial behaviors which can be exploited to find differences between positive and negative opinions~\cite{lucey2010automatically}.

\begin{figure}[!t]
\centering
\includegraphics[width=\textwidth]{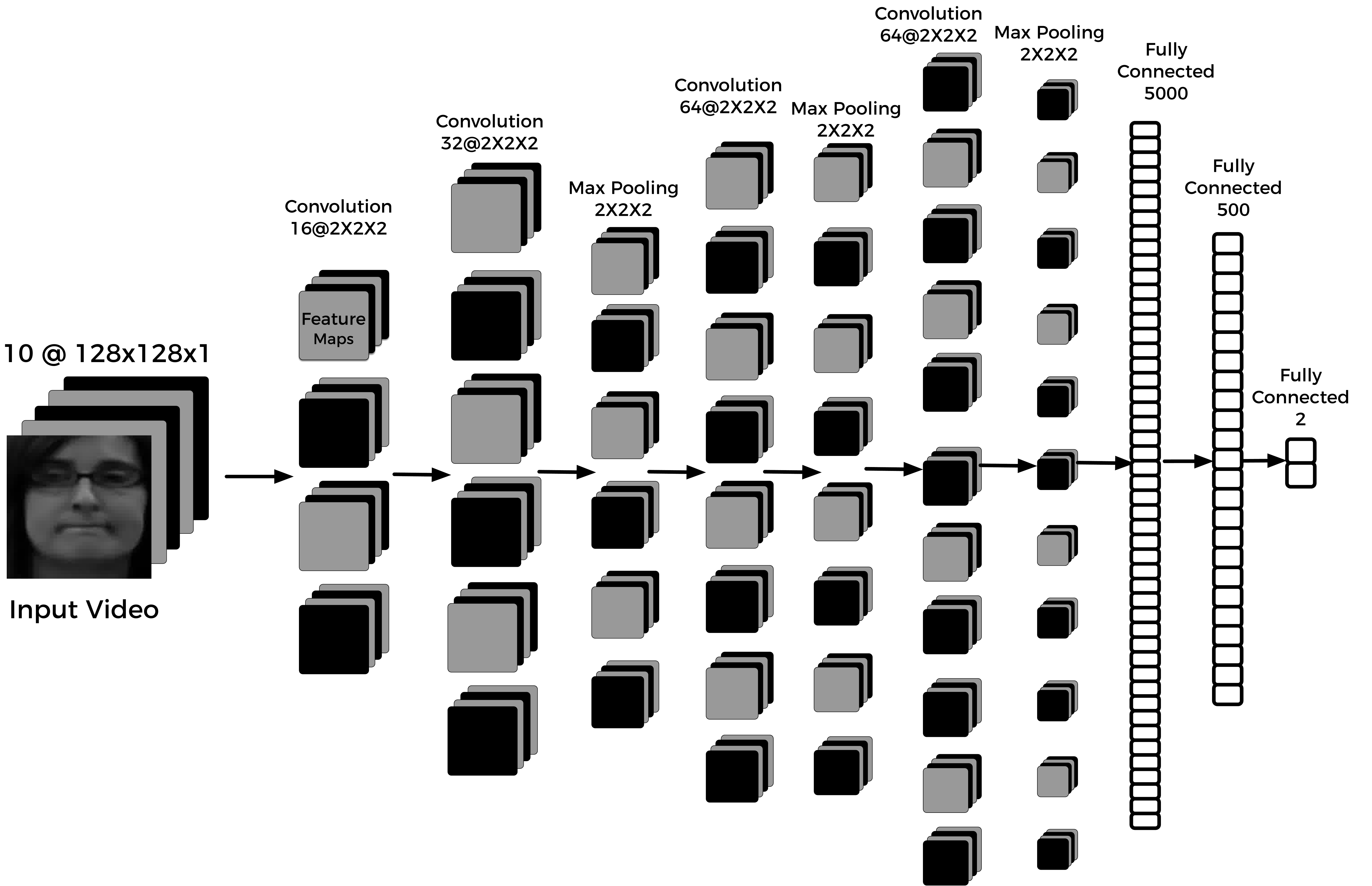}
\caption {3D-CNN based Visual Feature Extraction}
\label{fig:3dvisualcnn}
\end{figure}

The visual features are extracted from videos using a 3D CNN. The 3D-CNN contextually exploits both spatial and temporal patterns to accurately find the spatio-temporal association between a subjective and an objective utterance. In our experiments, the best results are obtained with a 9-layered 3D-CNN architecture as illustrated in Fig.~\ref{fig:3dvisualcnn}. The architecture details are presented in Table~\ref{table:3dcnn}.

\begin{table}[!t]
\caption{3D-CNN Architecture for visual cues feature extraction)}
\centering
\label{table:3dcnn}
\begin{tabular}{|l|l|l|l|}
\hline
\textbf{Layer} & \textbf{Type} & \textbf{Feature Map} & \textbf{Kernel} \\ \hline
1 & Convolutional3D & 16 & 2 x 2 x 2 \\ \hline
2 & Convolutional3D & 32 & 2 x 2 x 2 \\ \hline
3 & Max pooling3D & & 1 x 2 x 2 \\ \hline
4 & Convolutional3D & 64 & 2 x 2 x 2 \\ \hline
5 & Max Pooling3D & & 2 x 2 x 2 \\ \hline
6 & convolution3D & 64 & 2 x 2 x 2 \\ \hline
7 & Max pooling3D & & 1 x 2 x 2 \\ \hline
8 & Fully connected & 5000 & \\ \hline
9 & Fully connected & 500 & \\ \hline
10 & Fully connected & 2 & \\ \hline
\end{tabular}
\end{table}

\subsection{Multimodal Fusion}

In this section, we discuss multimodal fusion techniques. Multimodal fusion is the process of contextually integrating features collected from different modalities for the sentiment analysis task. In general, multimodal fusion techniques can be divided into early or feature-level fusion and late or decision-level fusion. Multimodal systems often outperform unimodal approaches as the correlation and discrepancies between modalities can help in achieving superior performance~\cite{d2015review}. 

\subsubsection{Feature-level (Early) Fusion}

In early fusion, first, the features are extracted from input modalities using either deep neural networks or state-of-the-art feature extraction techniques. The input features are concatenated and fed into a classifier. For example, audio features are extracted using OpenSMILE software, visual features are extracted using 3D-CNN and text features are extracted using BiLSTM. The features are then fused and fed into a shallow or deep learning classifier. The main advantage of feature-level fusion is that the cross-correlation between multiple modalities at an early stage helps in achieving better performance. On the other hand, the main disadvantage of early fusion is that the modalities must be tightly time synchronized, since incorrect time synchronization can lead to a poorly functioning system as the model will be unable to learn any cross-modal correlation~\cite{snoek2005early}.

\begin{figure}[!t]
\centering
\includegraphics[width=0.8\textwidth]{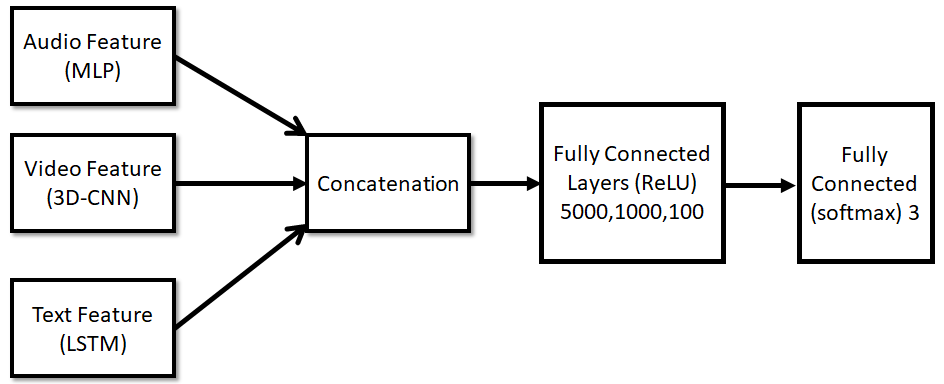}
\caption {Early (Feature-level) Fusion}
\label{fig:early}
\end{figure}

\subsection{Decision-Level (Late) Fusion}

In late or decision-level fusion, unimodal classifiers are used to identify the prediction for each modality. The local predictions are concatenated and further classified to achieve the final decision. The advantage of late fusion is that, the sampling rate at which the predictions are generated is the same, therefore the extracted predictions can be easily concatenated without up-sampling or down-sampling. However, the disadvantage is that the model cannot learn cross-modal correlations~\cite{natarajan2012multimodal}. The Fig \label{fig:late} displays the late fusion-level approach.  

\section{Persian Multimodal Dataset} \label{multi:multidataset}

In this section, we introduce the novel Persian multimodal dataset.

\subsection{Dataset Collection}

The YouTube website was used to collect Persian videos with a focus on product, movie and music reviews. The videos were found using keywords such as: \<nqd fylm> ("Movie critize"), \<m.h.swlAt AyrAny> ("Persian product"), \<nqd mwzyk> ("Critize music"). 

\begin{figure}[!t]
\centering
\includegraphics[width=0.8\textwidth]{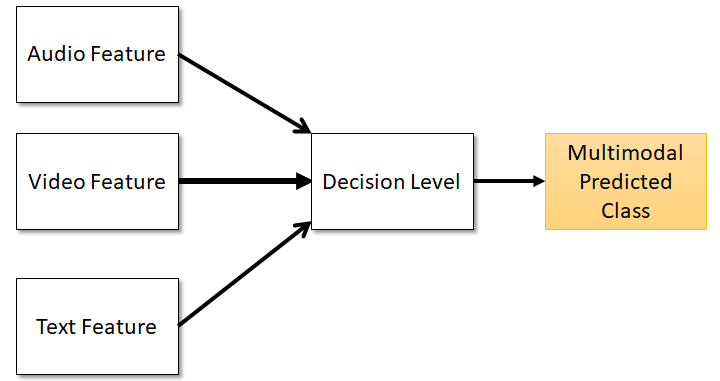}
\caption {Late (Decision-Level) Fusion}
\label{fig:late}
\end{figure}

\begin{figure}[!t]
\centering
\includegraphics[width=\textwidth]{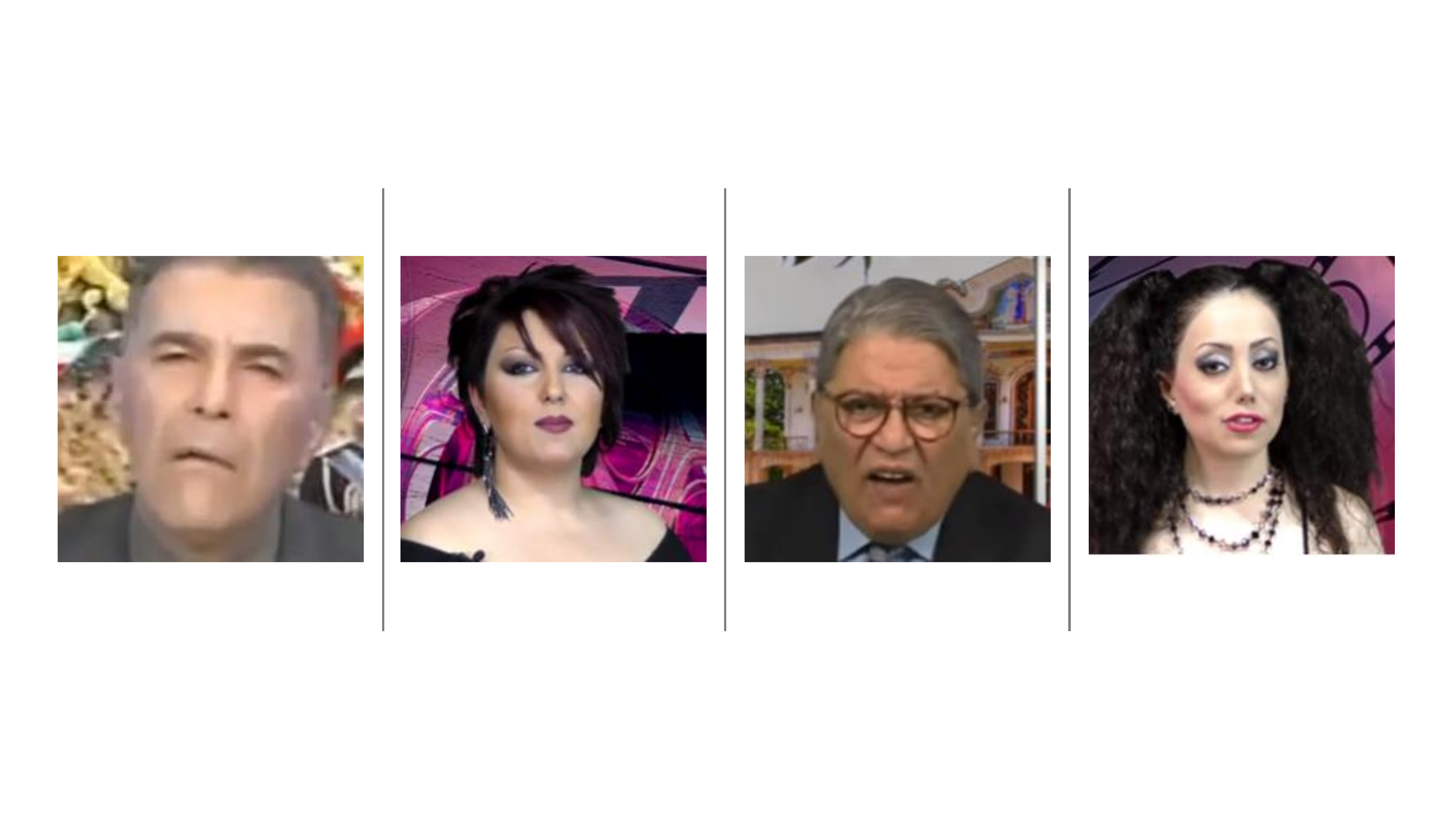}
\caption{Example snapshots of videos from our new
Persian multimodal dataset}
\label{fig:aud34}
\end{figure}

A total of 91 videos were selected that respected the following guidelines: 
\begin{itemize}
  \item The video must contain only one speaker
  \item The speaker should look directly in the camera
  \item The speaker face should be clearly visible
  \item There should not be any background noise or music in the recording
  \item The video should be recorded with high quality microphone and camera
\end{itemize}
While in the collected videos, the speaker has similar distance from camera, the background and lighting is variable between different videos. The length of the video is between 1-5 minutes. In addition, each video consists of 10--50 utterances. The dataset includes 15 male and 9 female speakers between 20 to 60 years age. Sample snapshots of our Persian multimodal dataset are shown in Fig.~\ref{fig:aud34}.

\subsection{Segmentation and Transcription}

All collected videos were transcribed manually with utterance start and end times. The transcription process comprised two stages. First, an expert transcriber manually transcribed all the videos. The transcribed text was reviewed by two other native speakers. In the second stage, the transcriptions were divided into utterances using pause details such as \<hm> \<oh> (hm, oh, etc.). 

Each video was segmented into an average ten utterances, resulting in a final dataset of 945 utterances with 754 subjective utterances and 191 objective utterances. Each utterance was then linked to audio and video streams as well as to manual transcription. The utterances have an average duration of 6 seconds. 

Three expert annotators categorized segmented utterances into subjective utterances (with expressed emotion) and objective utterances with no polarity (e.g., facts, figures etc.) category. The subjective segmentation was important to achieve fine-grained sentiment analysis. 
An utterance was categorized into the subjective category, if the sentence was carrying an opinion, belief, thought, feeling or emotion. Three rules were used to identify subjectivity in the sentence as outlined below: 

\begin{itemize}
 \item Explicitly criticising an entity. For example: \RL{yk fylm kmdy .tnz xwb kh lhjh hAy xwb bry mxA.tb} \RL{frAhm myknh}, "The comedy movie is really good and the viewers have good time when they are watching the movie".
 
 \item Referencing an opinion expressed by a third person. For example, \RL{ mntqdyn Az dydn fylm rA.dy nystnd}, "The movie critic are not satisfying with watching this film"
  
  \item Implicitly expressing a subjective opinion. \RL{mn py^snhAd nmyknm kh Ayn fylm bA xAnwAdh ngAh knyd}, "I am not recommended to watch this movie with family"
\end{itemize}

Detailed statistics of the dataset can be found in Table~\ref{fig:stastics}. 

\begin{table}[]
\caption{Persian Multimodal dataset statistics}
\centering
\label{fig:stastics}
\begin{tabular}{|l|l|}
\hline
Total number of positive segmented & 468 \\ \hline
Total number of negative segmented & 366 \\ \hline
Total number of subjective & 834 \\ \hline
Total number of objective & 180 \\ \hline
Total number of unique words in the dataset & 4065 \\ \hline
Total number of speakers & 24 \\ \hline
\end{tabular}
\end{table}

\subsection{Sentiment Polarity}

The utterances were annotated by three native Persian speakers (two male and one female) between 30 to 50 years old. All the speakers received a Master in Persian language. The annotators had three choices, positive (+1), neutral (0) and negative (-1). The polarity assignment included all three modalities (visual, audio and text). Table~\ref{example:tabl} shows examples of utterances obtained from one of the videos in the multimodal dataset, along with their translation and polarity. It can be observed that, a single video consists of both positive and negative utterances.

\begin{table}[!t]
\caption{Example utterances from the Persian Multimodal Sentiment Analysis Dataset}
\label{example:tabl}
\centering
\begin{tabular}{|l|l|l|}
\hline
\textbf{Persian Sentence} & \textbf{English Translation} & \textbf{Polarity} \\ \hline
\begin{tabular}[c]{@{}l@{}}\RL{b`d ^cnd sAl ^sAdmhr} \\ \RL{yk kAr xwb} \\ \RL{mnt^sr krd bh Asm} \\ \RL{bA tw `^sqm}\end{tabular} & \begin{tabular}[c]{@{}l@{}}After few years, \\ Shadmehr release good \\ work called "with \\ you my love\end{tabular} & 1 \\ \hline
\begin{tabular}[c]{@{}l@{}}\RL{bAzygr dwst dA^stny r.dA} \\ \RL{`.t'ArAn mA rw bh} \\ \RL{dydn fylm tr.gyb myknh}\end{tabular} & \begin{tabular}[c]{@{}l@{}}The good acting of \\ Reza Attaran help \\ to people watch the movie\end{tabular} & 1 \\ \hline
\begin{tabular}[c]{@{}l@{}}\RL{Ar^sAd q.sd} \\ \RL{mjwz bh fylm ndArh}\end{tabular} & \begin{tabular}[c]{@{}l@{}}The ministry of culture \\ did not give the \\ permission to the movie\end{tabular} & -1 \\ \hline
\begin{tabular}[c]{@{}l@{}}\RL{tw dwrAn phlwy yk mdt} \\ \RL{bh xA.tr trAnh} \\ \RL{syAsy zndAn bwd}\end{tabular} & \begin{tabular}[c]{@{}l@{}}During Pahlavi Dynasty \\ he was jailed \\ for few years\end{tabular} & -1 \\ \hline
\end{tabular}
\end{table}

Finally, manual gestures including smile, frown, head node and head shake were annotated manually to study the relation between words and gestures. The annotation was carried out by marking the utterances into these gestures. Expert coders manually annotated each utterance with gesture information. The average agreement of the gestures was 89.23\%.

\section{Experimental Results}\label{multi:experiment}

In this section, comparative experimental results for Persian multimodal dataset are discussed in detail. In this study, we focus on identifying effective strengths as compared to finding if an utterance is subjective or objective. Therefore, we removed all objective utterances for training the multimodal sentiment analysis framework. The features for text, audio and video modalities are experimentally determined and contextually utilized as described in the previous sections.

The results for text-based, audio-based and video-based unimodal sentiment analysis are summarized in Table~\ref{fig1:six3k}, Table~\ref{figfd:six4k} and Table~\ref{fig1:six5k} respectively. Experimental results show that the text-based classifiers achieved better accuracy as compared to audio-based and video-based unimodal sentiment analysis models. It is to be noted that the model is trained for utterance level sentiment analysis and not video level sentiment analysis. Generally, each video consists of 10-15 sentences. In addition, it is to be noted that there was no overlap of speakers in training, validation and test data for speaker independent analysis. Therefore, it can be concluded that the model generalizes well on unseen samples.

\begin{table}[!t]
\caption{Prediction Results: Text-Based unimodal Sentiment Analysis}
\centering
\label{fig1:six3k}
\begin{tabular}{|l|l|l|l|l|l|l|l|}
\hline
 & \multicolumn{4}{l|}{\textbf{Precision}} & \textbf{Recall} & \textbf{F-measure} & \textbf{Accuracy} \\ \hline
Positive & \multicolumn{4}{l|}{0.92} & 0.83 & 0.87 & \multirow{2}{*}{} \\ \cline{1-7}
Negative & \multicolumn{4}{l|}{0.88} & 0.94 & 0.91 & \\ \hline
Average & \multicolumn{4}{l|}{0.90} & 0.88 & 0.89 & 89.24 \\ \hline
\end{tabular}
\end{table}

Table~\ref{figfd:six4k} shows the experimental results for audio-based unimodal sentiment analysis. It can be seen that the positive features achieve better recall compared to negative features. However, the negative features achieve better F-measure as compared to positive features.

\begin{table}[!t]
\caption{Prediction Results: Audio-Based unimodal Sentiment Analysis}
\centering
\label{figfd:six4k}
\begin{tabular}{|l|l|l|l|l|l|l|l|}
\hline
 & \multicolumn{4}{l|}{\textbf{Precision}} & \textbf{Recall} & \textbf{F-measure} & \textbf{Accuracy} \\ \hline
Positive & \multicolumn{4}{l|}{0.78} & 0.84 & 0.81 & \multirow{2}{*}{} \\ \cline{1-7}
Negative & \multicolumn{4}{l|}{0.78} & 0.82 & 0.84 & \\ \hline
Average & \multicolumn{4}{l|}{0.78} & 0.83 & 0.82 & 82.79 \\ \hline
\end{tabular}
\end{table}

Table~\ref{fig1:six5k} displays the results for video-based unimodal sentiment analysis. As discussed earlier in section~\ref{multi:visual}, the 3D-CNN is used to extract features from videos. Experimental results show that the negative features achieve better precision and F-measure as compared to positive features. However, the positive features achieve better recall as compared to negative features. Experimental results show that the visual feature achieved lower performance as compared to acoustic and textual features. The main reason, the speakers have shown less emotion in the visual features as compared to other features.

\begin{table}[!t]
\caption{Prediction Results: Video-Based unimodal Sentiment Analysis}
\centering
\label{fig1:six5k}
\begin{tabular}{|l|l|l|l|l|l|l|l|}
\hline
 & \multicolumn{4}{l|}{\textbf{Precision}} & \textbf{Recall} & \textbf{F-measure} & \textbf{Accuracy} \\ \hline
Positive & \multicolumn{4}{l|}{0.76} & 0.85 & 0.80 & \multirow{2}{*}{} \\ \cline{1-7}
Negative & \multicolumn{4}{l|}{0.87} & 0.79 & 0.83 & \\ \hline
Average & \multicolumn{4}{l|}{0.81} & 0.82 & 0.81 & 81.72 \\ \hline
\end{tabular}
\end{table}

Table~\ref{tab:latefus} summarizes the results for late or decision-level fusion. As can be seen, comparative experimental results demonstrate that the full multimodal system comprising audio (A) + video (V) + text (T) modalities, and the part-multimodal system comprising V + T modalities achieve better accuracy as compared to other modality combinations. However, the A + T modality achieves better precision compared to other modalities. In addition, the V + T modality achieves better recall compared to the other modalities, whereas the full A + V + T modality achieves better F-measure compared to other modalities. Finally, the results show that the A + V modality achieves the least accuracy as compared to other multimodality combinations. As the experimental results show our proposed method is robust and achieves great performance even if one of the modalities such as visual, acoustic and textual features are not available. For example, the model achieved superior performance using A+V (audio and visual) without using T (textual).
The multimodal systems often exploit the correlation between modalities for predicting more accurate output as compared to unimodal systems. E.g. human facial expressions are often correlated with other modality such as their voice and the words which they are using  \cite{zuckerman1976encoding}.  

\begin{table}[]
\caption{Prediction Results: Late (decision-level) Fusion}
\centering
\label{tab:latefus}
\begin{tabular}{|l|l|l|l|l|l|l|l|l|}
\hline
\textbf{Modality}   & \textbf{} & \multicolumn{4}{l|}{\textbf{Precision}} & \textbf{Recall} & \textbf{F-measure} & \textbf{Accuracy} \\ \hline
\multirow{3}{*}{A+V}  & Positive & \multicolumn{4}{l|}{0.78}        & 0.79      & 0.79        & \multirow{2}{*}{} \\ \cline{2-8}
            & Negative & \multicolumn{4}{l|}{0.84}        & 0.83      & 0.83        &          \\ \cline{2-9} 
            & Average  & \multicolumn{4}{l|}{0.79}        & 0.78      & 0.78        & 81.18       \\ \hline
\multirow{3}{*}{V+T}  & Positive & \multicolumn{4}{l|}{0.89}        & 0.89      & 0.89        & \multirow{2}{*}{} \\ \cline{2-8}
            & Negative & \multicolumn{4}{l|}{0.91}        & 0.91      & 0.91        &          \\ \cline{2-9} 
            & Average  & \multicolumn{4}{l|}{0.88}        & 0.88      & 0.88        & 90.32       \\ \hline
\multirow{3}{*}{A+T}  & Positive & \multicolumn{4}{l|}{0.84}        & 0.93      & 0.88        & \multirow{2}{*}{} \\ \cline{2-8}
            & Negative & \multicolumn{4}{l|}{0.94}        & 0.87      & 0.90        &          \\ \cline{2-9} 
            & Average  & \multicolumn{4}{l|}{0.92}        & 0.84      & 0.88        & 89.24       \\ \hline
\multirow{3}{*}{A+V+T} & Positive & \multicolumn{4}{l|}{0.88}        & 0.90      & 0.89        & \multirow{2}{*}{} \\ \cline{2-8}
            & Negative & \multicolumn{4}{l|}{0.92}        & 0.90      & 0.92        &          \\ \cline{2-9} 
            & Average  & \multicolumn{4}{l|}{0.90}        & 0.87      & 0.89        & 90.32       \\ \hline
T-Only  & Average  & \multicolumn{4}{l|}{0.90}        & 0.88      & 0.89        & 89.24       \\ \hline
A-Only  & Average  & \multicolumn{4}{l|}{0.78}        & 0.83      & 0.82        & 82.79       \\ \hline
V-Only  & Average  & \multicolumn{4}{l|}{0.81}        & 0.82      & 0.81        & 81.72       \\ \hline
\end{tabular}
\end{table}

Table~\ref{tab:earlyfus} summarizes the results for early or feature-level fusion. Comparative experimental results show that the A + V + T and V + T modalities achieve better accuracy compared to other modalities. However, the A + T modality achieves better precision compared to other modalities. Furthermore, the V + T modality achieves better recall as compared to other modalities. Finally, the results show that the A + V modality combination achieves the least accuracy compared to other modalities.

\begin{table}[!t]
\caption{Prediction Results: Early (feature-level) Fusion}
\centering
\label{tab:earlyfus} 
\begin{tabular}{|l|l|l|l|l|l|l|l|l|}
\hline
\textbf{Modality} & & \multicolumn{4}{l|}{\textbf{Precision}} & \textbf{Recall} & \textbf{F-measure} & \textbf{Accuracy} \\ \hline
\multirow{3}{*}{A+V} & Positive & \multicolumn{4}{l|}{0.82} & 0.79 & 0.81 & \multirow{2}{*}{} \\ \cline{2-8}
 & Negative & \multicolumn{4}{l|}{0.84} & 0.87 & 0.85 & \\ \cline{2-9} 
 & Average & \multicolumn{4}{l|}{0.79} & 0.82 & 0.80 & 83.33 \\ \hline
\multirow{3}{*}{V+T} & Positive & \multicolumn{4}{l|}{0.90} & 0.89 & 0.89 & \multirow{2}{*}{} \\ \cline{2-8}
 & Negative & \multicolumn{4}{l|}{0.92} & 0.92 & 0.92 & \\ \cline{2-9} 
 & Average & \multicolumn{4}{l|}{088} & 0.90 & 0.89 & 90.86 \\ \hline
\multirow{3}{*}{A+T} & Positive & \multicolumn{4}{l|}{0.91} & 0.88 & 0.89 & \multirow{2}{*}{} \\ \cline{2-8}
 & Negative & \multicolumn{4}{l|}{0.91} & 0.93 & 0.92 & \\ \cline{2-9} 
 & Average & \multicolumn{4}{l|}{0.87} & 0.91 & 0.89 & 90.86 \\ \hline
\multirow{3}{*}{A+V+T} & Positive & \multicolumn{4}{l|}{0.85} & 0.98 & 0.91 & \multirow{2}{*}{} \\ \cline{2-8}
 & Negative & \multicolumn{4}{l|}{0.98} & 0.87 & 0.92 & \\ \cline{2-9} 
 & Average & \multicolumn{4}{l|}{0.97} & 0.84 & 0.90 & 91.39 \\ \hline
T-Only  & Average  & \multicolumn{4}{l|}{0.90}        & 0.88      & 0.89        & 89.24       \\ \hline
A-Only  & Average  & \multicolumn{4}{l|}{0.78}        & 0.83      & 0.82        & 82.79       \\ \hline
V-Only  & Average  & \multicolumn{4}{l|}{0.81}        & 0.82      & 0.81        & 81.72       \\ \hline
\end{tabular}
\end{table}

\begin{figure}[!t]
\centering
\includegraphics[width=\textwidth]{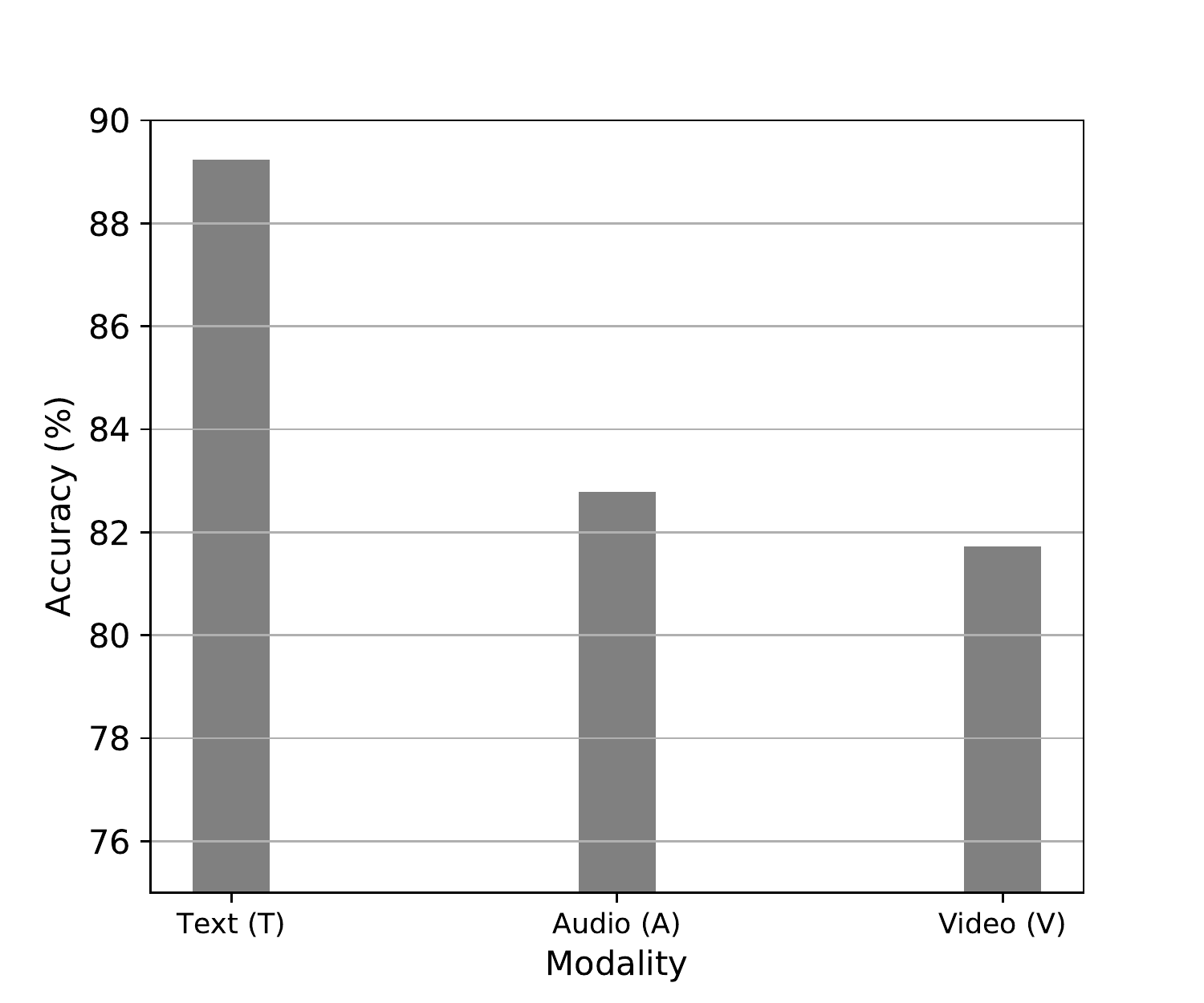}
\caption {Comparison of proposed Unimodal Persian Sentiment Analysis}
\label{fig:modality3453453}
\end{figure}

Fig.~\ref{fig:modality3453453} presents the accuracy of unimodal sentiment analysis models for text, audio and video modalities. It can be seen that the text modality achieves better accuracy compared to audio and video modalities.

\subsection{Discussion}

\begin{figure}[!t]
\centering
\includegraphics[width=\textwidth]{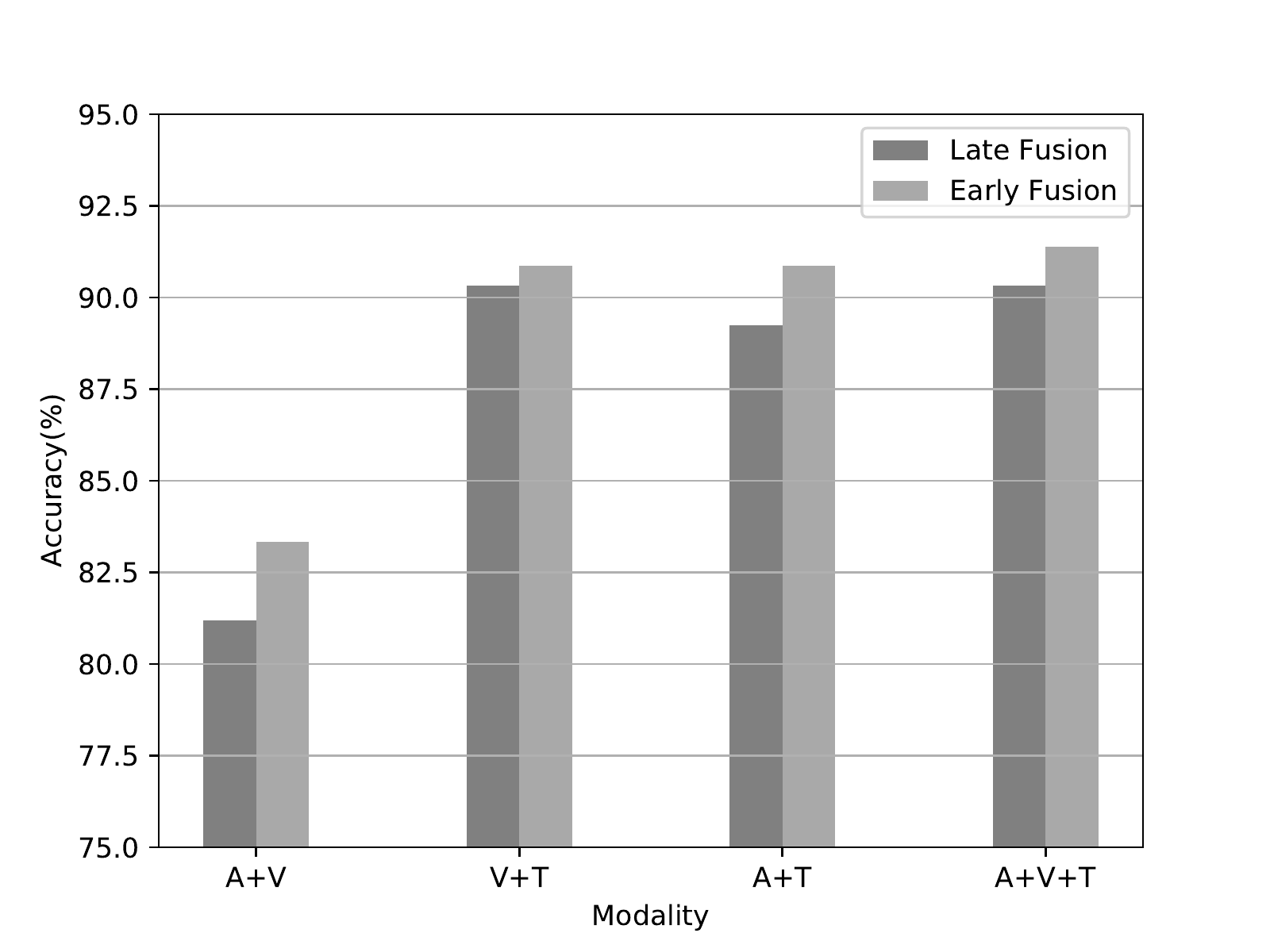}
\caption {Early (feature-level) vs Late (decision-level) Multimodal Fusion approaches: Comparison of prediction accuracy, precision, recall and F1-score}
\label{fig:latevseraly12}
\end{figure}

In the comparative experiments described above, all possible multimodal fusion combination including A + V, A + T, T + V, A + V + T were contextually considered. Fig.~\ref{fig:latevseraly12} presents the performance accuracy results of early and late fusion approaches. The results show that early fusion outperforms the late fusion strategy. Furthermore, the full A + V + T multimodal system with early (feature-level) fusion is seen to achieve the highest accuracy compared to late (decision-level) fusion and other early multimodal fusion combinations.  

In addition, the comparative experimental results show that multimodal fusion consistently achieves better performance compared to all unimodal sentiment analysis approaches. Amongst the latter, T-only analysis performs better than both other unimodal (V-only, A-only) approaches. The superior performance of the multi-modal (A+V+T) approach compared to the best unimodal (T-only) approach is attributed to the contextual fusion of acoustic, visual and textual features which are able to overcome the limitation of ambiguous words/phrases usage in Persian T-only unimodal sentiment analysis.

Finally, the main limitations of our proposed Persian multimodal sentiment analysis framework are identified below, which need addressed in future research:

\begin{itemize}
  \item The Persian multimodal sentiment analysis framework cannot detect subjective/objective utterances. For example, \RL{brym qsmthAy Az Ayn fylm bbynym w brgrdym}, "Let's go and see some part of the movie and come back". The proposed approach is unable to detect objective utterances. 
  \item The speakers in the videos are not speaking formally i.e. most of the videos comprise informal words to express the speaker's opinion which is difficult for the proposed approach to detect. For example, \RL{yAsr bxtyAry bA mwzyk rp^s .hrf dl jwAnhA rw myznh w xyly tA_tyr g_dr hst} "Yasr Bakhtiari with his rap music is saying what young people want and his music is really effective". 
  \item The current model is restricted to a single speaker and does not work in multi-speaker scenarios. For example, when the video consists of two speakers it cannot detect the polarity of sentences they are simultaneously expressing.
  \item The number of neutral discourses in the data is very less as compared to positive and negative cases. Therefore, we cannot train our proposed models on neutral comments as deep learning requires a large  training dataset. We intend to extend the presented multimodal dataset to include more objective discourse in order to facilitate multimodal subjectivity classification.
\end{itemize}

It is worth mentioning that the dataset was not translated into English before processing as the translation often fails to work on languages comprising lots of sarcastic and informal words \cite{lazard1992grammar}. Therefore, the multimodal approach was directly trained on Persian language without translating the input data into English.

\section{Conclusion and Future work} \label{multi:conclusion}

In text-based sentiment analysis, the usual source of information consists of n-gram, word-order, dependency relations and part-of-speech features that often prove inadequate to identify the overall polarity of the natural language sentence. On the other hand, videos comprise multiple modalities including text, audio and visual features which can be contextually exploited to enhance the polarity detection. In this paper, we presented a first of its kind multimodal dataset for Persian language, consisting of utterances and their sentiment polarity extracted from YouTube videos. In addition, a novel Persian multimodal sentiment analysis framework for contextually combining audio, visual and textual features was proposed. Our experimental results demonstrated that the fusion of Persian audio, visual and textual features outperform all other unimodal classifiers including text-based, audio-based and video-based sentiment analysis models. 

In future, we plan to address limitations with current unimodal and multimodal features through contextual extraction of more objective utterances using Persian dependency-rule based approaches~\cite{dashtipour2020hybrid}. Furthermore, we will explore multilingual approaches to detect polarity in multilingual videos, by exploiting our newly developed language-independent multimodal models~\cite{gogate2020cochleanet,gogate2020deep,gogate2020visual,gogate2019av} and multi-task learning approaches~\cite{xiong2018guided}.

\end{document}